# Hybrid Ensemble optimized algorithm based on Genetic Programming for imbalanced data classification

**Maliheh Roknizadeh,** *Faculty of Computer and Information Technology Engineering, Neyshabur Branch, Islamic Azad Universit, Neyshabur, Iran*
m.roknizade95@gmail.com

**Hossein Monshizadeh Naeen,** *Faculty of Computer and Information Technology Engineering, Neyshabur Branch, Islamic Azad University, Neyshabur, Iran*
monshizadeh@iau-neyshabur.ac.ir

**Abstract**

One of the most significant current discussions in the field of data mining is classifying imbalanced data. In recent years, several ways are proposed such as algorithm level (internal) approaches, data level (external) techniques, and cost-sensitive methods. Although extensive research has been carried out on imbalanced data classification, however, several unsolved challenges remain such as no attention to the importance of samples to balance, determine the appropriate number of classifiers, and no optimization of classifiers in the combination of classifiers. The purpose of this paper is to improve the efficiency of the ensemble method in the sampling of training data sets, especially in the minority class, and to determine better basic classifiers for combining classifiers than existing methods. We proposed a hybrid ensemble algorithm based on Genetic Programming (GP) for two classes of imbalanced data classification. In this study uses historical data from UCI Machine Learning Repository to assess minority classes in imbalanced datasets. The performance of our proposed algorithm is evaluated by Rapid-miner studio v.7.5. Experimental results show the performance of the proposed method on the specified data sets in the size of the training set shows 40% and 50% better accuracy than other dimensions of the minority class prediction.

**Keywords:** Bagging, Boosting, Hybrid ensemble method, Genetic programming algorithm, combination classifier, SMOTE

## 1. Introduction

The issue of unbalanced data processing has received considerable critical attention, because target issue is essential for a wide range of real world problems such as, fraud detection [1], oil- discovery [2], and medical diagnosis [3] . In unbalanced data, the number of samples of one class is usually much larger than the samples of the other class. A class with more data is called a majority class and a class with less data is called a minority class. In these issue, minority class ratios are often 1:100 or higher. In unbalanced data, the main challenge faced by many researchers is to correctly identify minority class samples. In the medical field, for example, the number of positive samples from a disease is much lower than the number of negative samples. While the importance of identifying samples related to the positive category is very high. The key role of class distribution in the design of a classifier and in standard data classification algorithms, balanced class distributions have received considerable attention in recent years. However, the use of these algorithms in unbalanced data classification can be not achieve acceptable results, because target algorithms the classification tends to the training samples of the majority class, which increases the number of errors in identifying positive samples. In recent years, a lot of high-quality research has been presented to solve addressing problem. The existing researches have shown that approaches based on classification combination have been significantly successful such as, SMOTEBOOST [4], Ada-BOOST [5], RUSBOOST [6], etc. Despite the high-quality research in combining the classifiers, there remain many challenges unresolved such as:

1- In existing classification hybrid algorithms, the focus is to create a trade-off between the majority and the minority class samples count. This trade-off is achieved either by over-sampling of minority class data points, or by under-sampling of majority class data points. By indiscriminately and blindly increasing the number of samples, it can lead to the production and expansion of some samples that do not play an affective role in the description. No attention to the importance of samples in data balancing is a weakness in previous studies and therefore has been one of the main challenges in our study.
2- To the best of our knowledge, previous studies have suffered from a paucity of considering the optimal basic classifiers number.
3- In most previous studies, the classifier voting is usually weighted and based on its decision strength on the data sets. So far, setting out these weights to increase the general decision strength of the combined classification has not been considered [5].

In this paper, two algorithms have been used to improve the sampling (first challenge) of the training data set. Generally, the purpose of this paper is to improve the efficiency of the ensemble method in the sampling of training data sets, especially in the minority class, and to determine better basic classifiers for combining classifiers than existing methods.

The rest of this paper is organized as follows. Section 2 presents related work. In section 3 gives details of proposed algorithm. Section 4 summarizes the results and discussion followed by the conclusion in section 5.

## 2. Related work

One of the greatest challenges is to improve the performance of classifiers trained with unbalanced datasets in the field of machine learning. A large and growing body of literature has investigated to the problem of data unbalance. In order to address the problem in the dataset, researchers have proposed many methods [7], divided these approaches into three categories: data-level methods, algorithm-level methods and hybrid methods.

- **The data-level methods**

    Common data mining algorithms usually perform poorly in the face of the problem of unbalanced class, because by not considering data samples in the minority class, those increase overall accuracy. However, the samples in the minority class are more important in many applications than the samples in the other class, [8-11] classifying unbalanced data, due to the high importance for recognizing minority class samples, correct classification is very important [12, 13]. Previous research has shown that increasing the sample size reduces the error rate of unbalanced classification [14].This issue was also confirmed by [15], that presented similar results using fuzzy classification. [14]have presented that there is a significant relationship between class conflict and imbalance in unbalanced data classification, although the level is not defined. [16]many research has investigated the separable class [17], demonstrated that class conflict, such as unbalanced class distribution, is an obstacle to classification performance.

- **The algorithm-level methods**

    Later researchers have proposed many other algorithm-level methods on improved SVM algorithm such as Z-SVM [18] and GSVM-RU [19]. The Z-SVM uses the Z parameter for a positive metamorphic motion to maximize the G-mean value, while the GSVM-RU uses the grain-in calculations to display aggregated information to improve classification performance.

- **The hybrid methods**

    Ada-Boost [20], Bagging [21], and Random-forest [22] are popular methods based on a combination of classifiers. In [5] has been comprehensively investigated on ensemble learning techniques in binary class problems. Experimental comparisons and various analyzes of the ensemble algorithms of many methods performed such as, 1) classic ensemble methods like, Ada-Boost, Ada-Boost M1, Bagging, 2) Boosting cost sensitive like Ada-C2, 3) Boosting based on ensemble method like, RUS-Boost, SMOTE-Boost, MSMOTE-Boost, 4) Bagging based on ensemble method like, Under-Bagging, Over-Bagging, MSMOTE-Bagging, 5) the hybrid ensemble methods like, Easy-Ensemble [23], Balance-Cascade [24] on the 44 UCI machine learning datasets. The AUC results showed that RUS-Boost, SMOTE-Bagging and Under-Bagging returned an improve classification compared to other ensemble algorithms, especially RUS-Boost, which had the least computational complexity among other methods. A review of [25] also noted that no attempt has been made to understand the diversity of classification errors.

3. **Method**

As described in the introduction, existing unbalanced data classification methods such as, data-level (internal level), algorithm-level (external level) and cost-sensitive methods. Despite the methods presented in each categories and the valuable research, only, approaches based on the combination of classifiers have been significantly successful. However, some challenges remain unresolved.

A. **The proposed method**

This paper has been proposed an approach to the challenges. The flowchart outline of the proposed approach is shown in Figure 1.

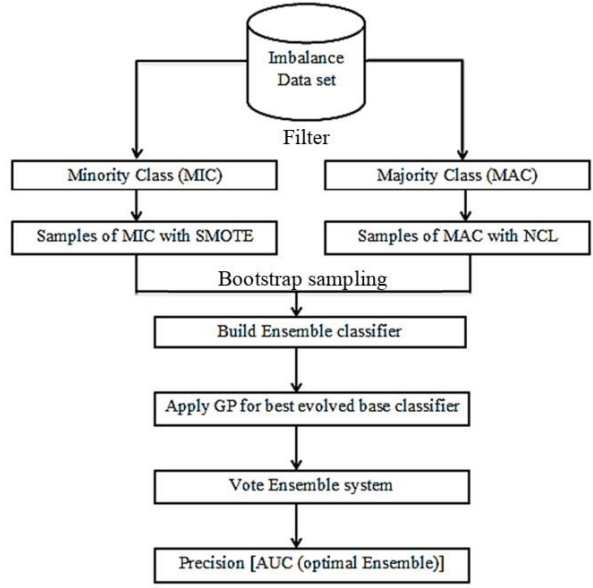

**Figure1.** The flowchart of the proposed approach

According to the flowchart approach proposed in Figure 1 in the Bagging process, the training data set is divided into two parts, based on the minority class (positive) and the majority class (negative). And then a good number of both classes enter the GP-Boost classifier. Asymmetric Bagging is possible to use as many unbalanced data sets as possible. Non-uniform separation of the original data and the bootstrap method can ensure the diversity of the training subset. For each training subset, unbalanced data sets, over-sampling (random sampling / SMOTE) and under-sampling (random sampling / NCL[1]) were used to balance the distribution of minority and majority classes. To reduce the variance and bias of the basic Ada-Boost learner, this model optimizes and completes within each sub-classifier by the Genetic Programming Algorithm (GP). The final output is better generated by using the voting combination of each sub-classifier.

**1) classifying using a combining of classifiers**

In this study, due to the high accuracy of predicting the composition of classifiers, the use of this method for data classification was considered. Using this method, the samples and base classifier better defined in the previous steps are considered as input to this classifier. It should mention that the algorithm for classifying the Ada-boost method and the basic algorithm used in all classifiers is the tree method. After initializing the sample weight distribution, the algorithm is executed to create a number of better classifiers. A new classifier in each performance is Create using training samples and weight distribution. Then the amount of error obtained by this classifier is calculated using Eq.1 on the training set.

$$\epsilon_j^t = \frac{1}{2} \sum_{(i,y) in B} w_{i,y}^t \cdot \left(1 - h_{j,t}(x_i, y_i) + h_{j,t}(x_i, y_i)\right) \qquad (1)$$

Determining the weight of the classifier is obtained by considering the error caused by the same classifier in Eq.2.

$$\beta_j^t = \frac{\epsilon_j^t}{(1 - \epsilon_j^t)} \qquad (2)$$

---

[1] Negative Correlation Learning

In updating the sample weight distribution for later execution, if the sample label is predicted correctly, the weight of the same sample in the next constant distribution will otherwise increase by $\beta_j^t$ [8]. After updating the sample weight distribution, the process of normalizing this distribution is performed. Then the next run to create a new classifier is done in the same way. This process of creating a classifier continues until the specified number is reached. After the algorithm is executed, the weights assigned to each classifier are used to determine the label of the given sample class. For example, in the classification of two classes, two variables are considered, one for positive samples and the other for negative samples. If the classifier predicts the sample label positively, then the weight of that classifier is added to the relevant variable. This is done for all classifiers, and at the end, the sample label receives the maximum weight from that class.

2) **Optimization of base classifiers using genetic programming algorithm**

To achieve maximum efficiency in combining classifiers, we need to optimize the basic classifiers produced in the training phase for each training subset. For this purpose, in this study, the GP algorithm was considered [26]. Therefore, this algorithm labels the classes specified for the training samples by each classifier with the weights related to each classifier and some parameters of the GP algorithm such as number of generations, number of the initial population, genius selection rate, and composition rate. And tree mutations are considered as input to the GP algorithm. The first step in the GP algorithm is to define the tree structure, the initial population and to define the appropriate fitting function according to the problem. Then the next step is how to select the parent to create the next generation using the selection operator and generate the next generation using the combination and mutation operator, and the last step is to determine the number of generations to achieve the optimal answer. In the following, each of the steps of the GP algorithm will be examined by the proposed method. The module (steps 2 to step 8) is repeated from the first $p_1$ subset to the last $p_T$ subset.

**Step1**. The index t is set from one. Each training pattern is assigned the same weight, $w_i = 1$ for i =1, …, N, where N is the total number of training data.

**Step2**. It is possible to create samples of training i in the $TR_t$ training set is $p_i = \frac{w_i}{\sum w_i}$ which is the sum of all members of the training set. Select n samples by placement the training set.

**Step3**. GP execution for classifiers in the $p_t$ subset with the $TR_t$ training set in the previous step builds the best derive $h_t: x \to y$ hypothesis tree.

**Step4.** Each member of the $TR_t$ training set passes through this $h_t$ tree to predictions $y_i^{(p)}(x_i)$ for i=1, …, N.

**Step5.** Calculate error $L_i = L\left(\left|y_i^{(p)}(x_i) - y_i\right|\right)$ for each training sample. The L error function is defined in the next steps.

**Step6.** Calculate the average error using: $\bar{L} = \sum_{i=1}^{N} l_i p_i$

**Step7**. Calculate the confidence level in the predictor by $\beta = \frac{\bar{L}}{1-\bar{L}}$.

**Step8**. Update weights using $w_i := w_i^{1-\bar{L}}$.

**Step.9.** t = t +1 If t ≤ T, then go to step 2.

**Step10.** For a unique $x_i$ input in the experimental data, a T tree is obtained. The best classifiers obtained from the sub-populations make a $h_t$ prediction for t = 1,..., T. The total prediction $h_f$ is obtained using Eq.3 using the predictor T.

$$h_f = \min\{ y \in Y: \sum_{t:h_t \leq y} \log(\tfrac{1}{\beta_t}) \geq \sum_t \log(\tfrac{1}{\beta_t})\} \tag{3}$$

In **step 5.** The error function L has a range in L ∈ [0, 1]. The error function used is given in Eq.4.

$$L_i = \frac{\left|y_i^{(p)}(x_i) - y_i\right|}{max_{i=1,\dots,N}\left|y_i^{(p)}(x_i) - y_i\right|} \tag{4}$$

β is the confidence level in the predictor, meaning that the lower the β value, the higher the confidence.

**Step9.** The smaller the error, the lower the weight, which makes the model less likely to be selected as a member of the training set for the next sub-population.

**Step 10** offers the middle weight. Assume that each better tree $h_t$ has a predictor $y_i^{(t)}$ on i pattern and a value of $\beta_t$ is assigned and the predictions are re-labeled for i-pattern, in Eq.5 has been shown.

$$y_i^{(1)} < y_i^{(2)} < \cdots < y_i^{(T)} \tag{5}$$

Then the sum $\log(\tfrac{1}{\beta_t})$ when t reaches its smallest value, thus inequality is established. The prediction of the better tree t is accepted as the final prediction. If all $\beta_t s$ are equal, the median is equal to the same.

4. **Experiments**

This section details the implementation of the proposed method.

➢ **Unbalanced data sets used in experiments**

This paper focuses on the problem of classifying unbalanced data sets with two classes, that positive samples are called the minority class and the negative samples are the majority class. The UCI Machine Learning Repository dataset is used. The size and distribution of the minority (positive) and majority (negative) classes related to this data set are described in Table 1.

Table 1.characteristic of imbalanced datasets

| Name | description | Total | Minority Num | Class % | IR | Feat. # Type |
|---|---|---|---|---|---|---|
| Ion | Ionosphere radar signal[103] | 351 | 126 | 35.8% | 1:3 | 34R |
| $Yst_1$ | Yeast protein* (mit) [103] | 1482 | 244 | 16.5% | 1:6 | 8R, Z |
| $Yst_2$ | Yeast protein* (me3) [103] | 1482 | 163 | 10.9% | 1:9 | 8R, Z |
| Vow | Vowel* (class0) [104] | 988 | 90 | 9.1% | 1:10 | 13R |

The experiments used six unbalanced datasets with a binary classification problem, which are described in Table 1. The datasets marked with * were multi-class classification problems that were sub-divided into binary classifications, one main class (minority) and one other class as the majority class. Attribute types are real (R) and integer (Z). In selecting the data set in terms of different levels of difficulty, the problem of imbalance, dimensions, size and types (characteristics) have been carefully considered. Minority class intervals are between 9% and 35% of the total sample. This data set does not include missing values. The tools used in this study are the tools used in the implementation of existing studies. How to get acquainted with them has been through authoritative papers and pages of their authors. The proposed method has been tested on a system with a Core i5 processor with GB4 memory in Windows 7. Rapid-miner Studio v.7.5 software has been used to implement the components of the proposed method.

> **The results proposed method Evaluation**

The main purpose of the experiments is as follows:
- o Investigate the impact of training set size on the error of the proposed method, implementation of the proposed method on the training set with the size of 20%, 30%, 40%, 60%, 80%, and 90%.
- o Investigate the accuracy of predicting minority class (positive) with the proposed method and methods compared, the effect of training sample size on the accuracy of minority class (positive).

The parameters used in the experiments are shown in Table 2. In the proposed method, the population size is 200 and the number of generations is 30. The number of repetitions was set to 10 Boosting techniques.

Table2. The parameters of experiment

| Name | Value |
|---|---|
| Number of subpopulation | 40 |
| Size of population | 200 |
| Max_depth_for_new_trees | 6 |
| Max_depth_after_crossover | 17 |
| Max_mutant_depth | 4 |
| Grow_method | Grow |
| Select_method | Tournament |
| Tournament_k | 6 |
| Crossover_func_pt_fraction | 0.7 |
| Crossover_any_pt_fraction | 0.1 |
| Fitness_prop_repro_fraction | 0.1 |
| Parsimony_factor | 0 |

According to the results obtained by the proposed method, the overall predict accuracy, and the predicted accuracy of the minority class with a training data set in the sizes of 20% to 90%, as well as the evaluation criteria has investigated. Figure 2 shows the effect of training sample size on minority class accuracy.

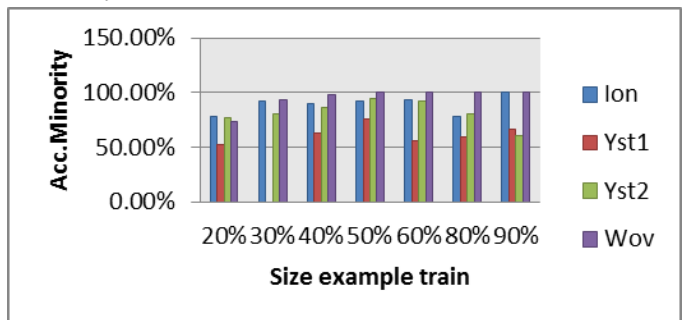

**Figure2**. Accuracy Minority

The diagram shows that, on average, the performance of the proposed method on the specified data sets in the size of the training set shows 40% and 50% better accuracy than other dimensions of the minority class. The following are diagrams of the impact of training sample size on an error and AUC evaluation criteria in Figure (3, 4) and the results of the proposed method have been improved in 40% and 50% samples.

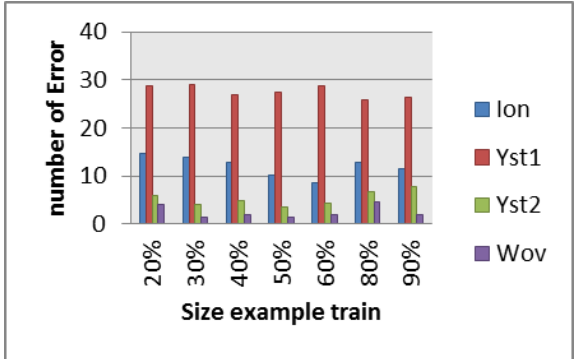

**Figure 3**.The number of error

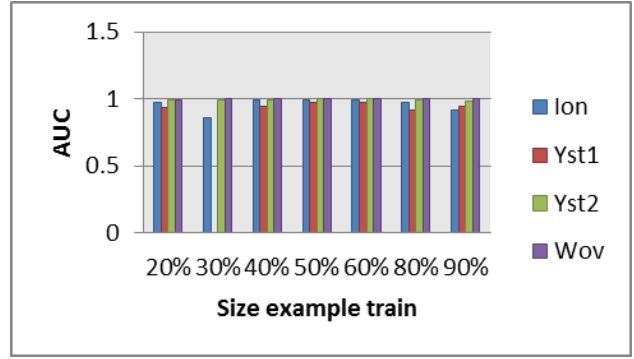

**Figure 4**.AUC evaluation

> **the Comparison proposed method with other algorithms**

In this section, we compare the results of the proposed method with the combined algorithms in the existing studies in unbalanced data sets.

**Table3**.The comparison proposed method

| Data set | EBag-GP Acc [27] | | EBoost-NB[27] | | EBoost-SVM[27] | | **E(B&BGP)** | |
|---|---|---|---|---|---|---|---|---|
| | Minority | Majority | Minority | Majority | Minority | Majority | **Minority** | **Majority** |
| Ion | 73.8% | 95.3% | 63.4% | 88.9% | 87.5% | 99.1% | **92.45%** | **88.52%** |
| Yst1 | 40.8% | 94.6% | 43.4% | 96.4% | 32.8% | 97.4% | **76.19%** | **72.40%** |
| Yst2 | 64.0% | 97.4% | 66.7% | 98.0% | 58.0% | 97.9% | **94.74%** | **96.64%** |
| Wov | 74.3% | 92.4% | 87.5% | 91.8% | 25.9% | 80.9% | **93.8%** | **98.46%** |

Table 3 and Figure (5, 6) shows the accuracy results obtained from the proposed method and GP, NB, and SVM methods. Experiments are performed by dividing equally (50% 50%) the training and testing data. The results show that the accuracy of the (positive) minority class has improved compared to the other three methods.

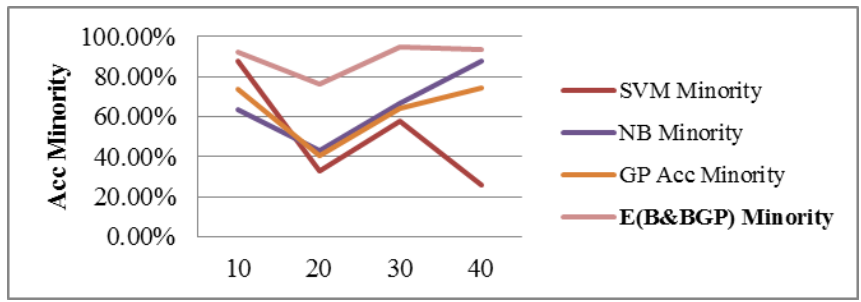

**Figure5**.Accuracy of Minority class prediction

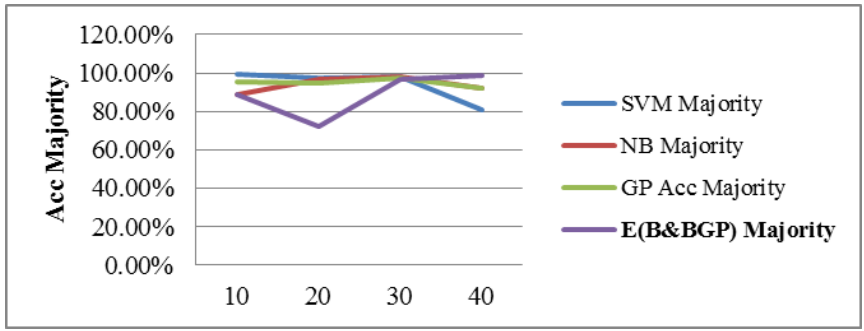

**Figure6**.Accuracy of Majority class prediction

Table 4 and Figure 7 shows the performance of the AUC evaluation criterion for the proposed method at different rates of imbalance in the unbalanced data set is better than the methods compared.

**Table4**.AUC evaluation of the proposed method

| Data-set | IR | Cost-sen[5] | | Boosting-based[5] | | | | | | Hybrids | | E(B&BGP) |
|---|---|---|---|---|---|---|---|---|---|---|---|---|
| | | C21 | C24 | RUS1 | RUS4 | SBO1 | SBO4 | MBO1 | MBO4 | EASY | BAL | |
| Yst1 | 2.46 | .660 | .675 | .718 | .699 | .700 | .698 | .713 | .708 | .714 | .683 | **.972** |
| Yst2 | 9.08 | .917 | .920 | .933 | .924 | .896 | .899 | .881 | .881 | .941 | .929 | **.999** |
| Wov | 10.10 | .970 | .970 | .942 | .967 | .989 | .991 | .952 | .952 | .941 | .938 | **1.00** |
| Shut | 13.87 | .999 | .999 | 1.00 | 1.00 | 1.00 | 1.00 | 1.00 | 1.00 | 1.00 | 1.00 | **1.00** |

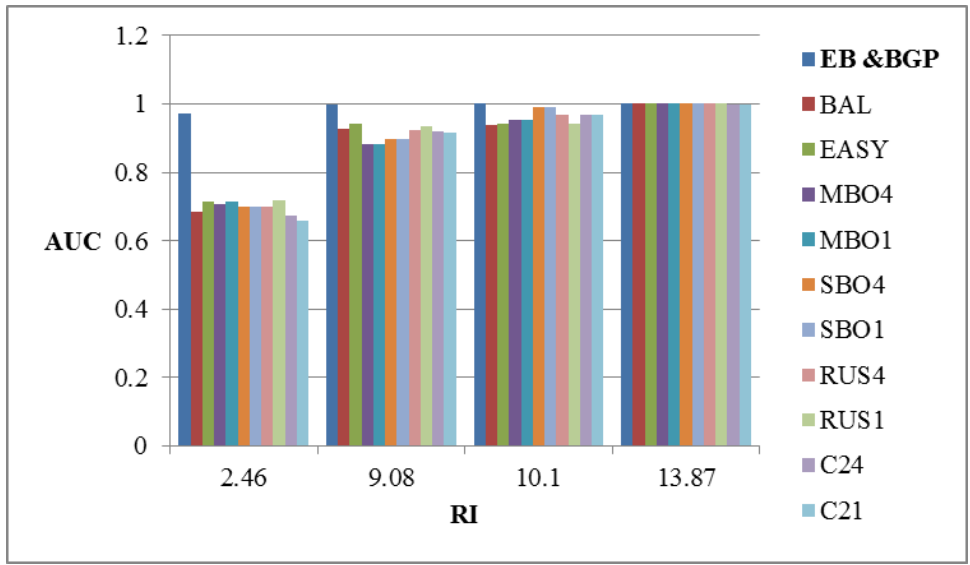

**Figure7**.AUC evaluation of proposed method

## 5. Conclusion

There are two primary aims of this study: 1.To investigate develop an ensemble learning method for classifying unbalanced datasets using (GP) 2.To ascertain a new selected ensemble method to modify the ensemble method using GP. This study, using GP for the selected ensemble method, presented a single combined solution of GP for the existing ensemble method. The two main innovations of the proposed method are such as, the use of centralized evolutionary selection to find small groups of high-participation individuals for the ensemble and also finding a diverse set of GP functions that skillfully controls the output of members to determine which ensemble is the final decision of the classification. The modified ensemble method was performed on 4 unbalanced datasets and results were confirmed in comparison with selected existing ensemble algorithms such as, Bagging and Boosting techniques; Standard GP, NB and SVM. The results showed that without using the proposed method, the ensemble organized is vulnerable to bias learning (due to the trend to one class that usually is the majority class).

## Acknowledgment

This article is taken from my master thesis. Thanks to my supervisor Dr.Monshizadeh Naeen for his help and guidance.

## Preferences